\documentclass[10pt,twocolumn,letterpaper]{article}

\usepackage{iccv}
\usepackage{times}
\usepackage{epsfig}
\usepackage{booktabs}
\usepackage{graphicx}
\usepackage{amsmath}
\usepackage{amssymb}

\usepackage[breaklinks=true,bookmarks=false]{hyperref}

\iccvfinalcopy % *** Uncomment this line for the final submission

\def\iccvPaperID{4280} % *** Enter the ICCV Paper ID here

\ificcvfinal\fi

\begin{document}

%%%%%%%%% TITLE
\title{HybridGazeNet $:$ Geometric model guided Convolutional Neural Networks for gaze estimation}

\author{ Shaobo Guo\\
Horizon Robotics\\
{\tt\small shaobo92@mail.ustc.edu.cn}
\and
Xiao Jiang\\
Horizon Robotics\\
{\tt\small xiao01.jiang@horizon.ai}

\and
Zhizhong Su\\
Horizon Robotics\\
{\tt\small zhizhong.su@horizon.ai}

\and
Rui Wu\\
Horizon Robotics\\
{\tt\small rui.wu@horizon.ai}

\and
Xin Wang\\
Horizon Robotics\\
{\tt\small xin.wang@horizon.ai}
}

\maketitle
\ificcvfinal\thispagestyle{empty}\fi

%%%%%%%%% ABSTRACT
\begin{abstract}
   As a critical cue for understanding human intention, human gaze provides a key signal for Human-Computer Interaction(HCI) applications. Appearance-based gaze estimation, which directly regresses the gaze vector from eye images, has made great progress recently based on Convolutional Neural Networks(ConvNets) architecture and open source large-scale gaze datasets. However, encoding model-based knowledge into CNN model to further improve the gaze estimation performance remains a topic that needs to be explored. In this paper, we propose HybridGazeNet(HGN), a unified framework that encodes the geometric eyeball model into the appearance-based CNN architecture explicitly. Composed of a multi-branch network and an uncertainty module, HybridGazeNet is trained using a hyridized strategy. Experiments on multiple challenging gaze datasets shows that HybridGazeNet has better accuracy and generalization ability compared with existing SOTA methods. The code will be released later.
\end{abstract}

%%%%%%%%% BODY TEXT
\section{Introduction}

\label{intro}
Human Gaze plays an important role in understanding human attention. Gaze estimation task has a long history in the domain of biometrics, and has various applications in different human computer interaction (HCI)\cite{chuang2014estimating} areas, such as virtual reality application (VR)\cite{piumsomboon2017exploring} and driver monitor system (DMS)\cite{chuang2014estimating} \cite{ji2002real}. 
\begin{figure}[ht]
	\centering
	\includegraphics[width=8cm]{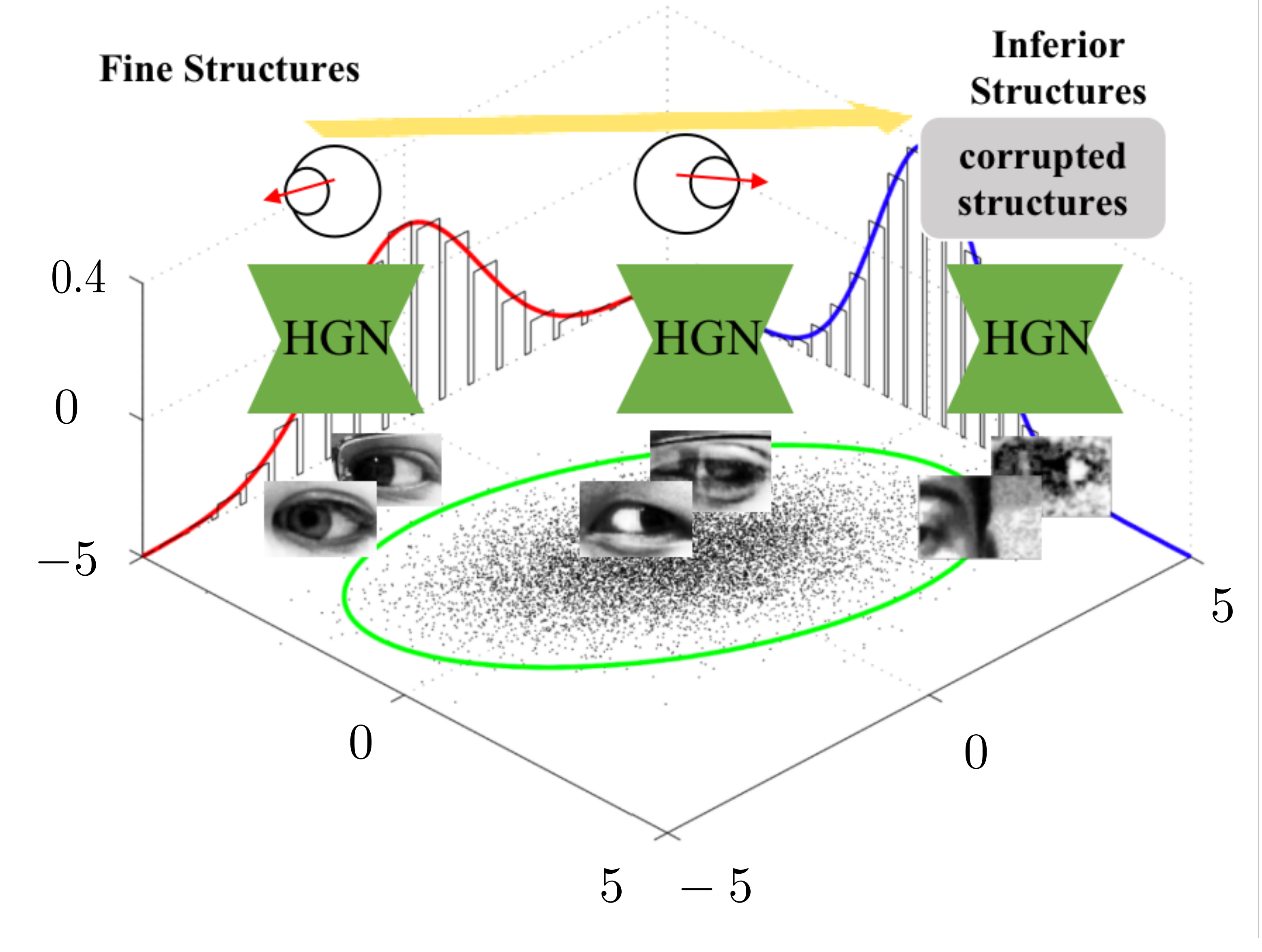}
	\caption{Framework of the proposed method. HybridGazeNet(HGN) is a structure aware model which takes 2D eye images as input, predicts gaze direction, eye landmarks, and eye image quality. The distribution of predict quality on each dataset is approximately Gaussian distribution.}
	\label{fig:overview}
\end{figure}

As a non-verbal cue, human gaze direction is a ray emitted from the center of the eyeball, and the gaze estimation task is to estimate gaze directions based on image data\cite{guestrin2006general}\cite{hansen2009eye}\cite{kar2017review}. Gaze estimation methods can be generally classified into geometric-based methods (GBMs) and appearance-based methods (ABMs). GBMs\cite{alberto2014geometric}\cite{villanueva2013hybrid}\cite{jianfeng2014eye} rely on 3D eyeball models to estimate gaze direction, such as pupil center corneal reflections(PC-CR)\cite{guestrin2006general} model and pupil center eye corner(PC-EC)\cite{sesma2012evaluation} model. GBMs require specialized hardware to capture detailed eye region images in order to build accurate 3D eyeball model, which limits its application scenarios. With the rapid development of deep learning theory\cite{krizhevsky2017imagenet} and the release of several large scale gaze datasets\cite{zhang2017mpiigaze}, appearance-based methods become popular in this gaze estimation area. ABMs\cite{zhang2015appearance}\cite{zhang2017s}\cite{zhang2018revisiting}\cite{xiong2019mixed}\cite{cheng2018appearance}\cite{cheng2020coarse} usually regress gaze directions from input images with carefully designed Convolution Neural Networks(ConvNets) in an end-to-end manner. However, data-driven methods need a large amount of qualified labeled samples which are quite expensive to collect. So here is the dilemma - On the one hand, GBMs could achieve decent performance with manually designed system and perform better over different image domains but they have strict requirements for image-sensors. On the other hand, ABMs are easy to train and deploy, but they depend heavily on high-quality training data.

Hence recently, a rising trend is to combine the advantages of above methods. In \cite{park2018learning}, the authors trained eye landmark localization model on synthetic data \cite{wood2016learning} and then take eye landmark localization as input for GBMs in a two-stage manner. And in \cite{Park_2018_ECCV}\cite{yu2018deep}\cite{zhu2017monocular}, the authors added geometric information, such as head pose or eye shape as regularization during training appearance-based ConvNets model. However, such regularization can not provide explicit geometric structure of gaze model. 

% briefly introduce the implementation of HGN
In this paper, we introduce an end-to-end framework called HybridGazeNet(HGN) which integrates a gaze model into a ConvNets, and can be trained using a hybrid strategy. After analyzing the relation between gaze directions and eyeball landmark locations,we find it's possible to extract the learnable parameter from the gaze model as ConvNets output, and therefore we can encode the geometric regularization into the ConvNets model. The construction process of HGN will be introduced in Section \ref{hgn}. After that, we propose a novel strategy to train this model. Since the off-the-shelf gaze datasets don't provide the geometric information of eyeball model, we use UnityEyes\cite{wood2016learning} to generate synthetic data, and train the geometric parameter with synthetic data supervision while the gaze supervision comes from the real data. However, training on both real and synthetic data simultaneously would bring out other issues. Theoretically, we figure out that there're three misalignments need to be fixed during hybrid training, these will be declared in Section \ref{training}. 

Moreover, unlike traditional computer vision tasks such as image classification\cite{deng2009imagenet} and object detection\cite{lin2014microsoft}, training data can be labeled by human. However, gaze samples are automatically recorded with carefully designed equipment. Even so, the quality of gaze samples is hard to control during the data recording procedure. Blinks, self occlusions and human distractions will degrade the data quality. In our work, in order to acquire precise .with the geometric aware ability of HGN, we further introduce a simple but effective uncertainty module to describe the quality of gaze samples. All the factors caused by the blinks, failed landmark detection etc. would damage the eye structure of input images. 
Since HGN is sensitive to the eye structure, we introduce an uncertainty module into it, which can alleviate the negative effect caused by inferior labeled gaze data during training procedure. As a compensate module of HGN framework, this would enable the model with dynamic adjustment ability over different quality of data and improve the robustness of the HGN framework.

Experiment results shows, with the end-to-end designation, HGN could benefit from both sides of ABMs and GBMs while avoiding the restriction of each method. Different with GBM methods, HGN is more simple and robust to various quality level of input data. Equipped with the regularization of explicit gaze model, the generalization of HGN outperforms former methods and make the gaze estimation framework more interpretable

\begin{figure*}[ht]
	\centering
	\includegraphics[width=18cm]{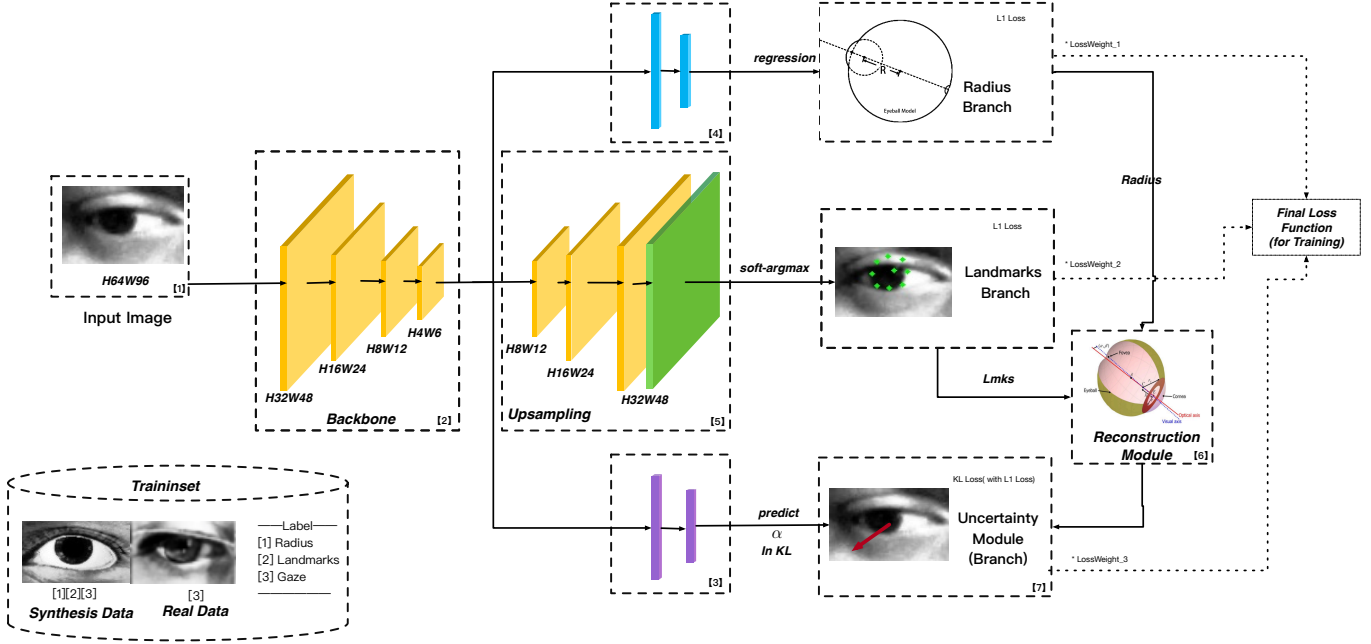}
	\caption{HGN network structure. We use hybrid data(i.e. real data, and synthetic data) for training. There are three branches: landmark prediction head, eyeball radius head, and uncertainty head. With the prediction of eyeball radius and eye landmark, gaze direction can be calculated through reconstruction module. Also a uncertainty module is proposed to evaluate the quality of input samples. Modules: 1.Data input; 2.Backbone; 3.Uncertainty head; 4.Eyeball radius head; 5.Eye landmarks decoder; 6.Reconstruction module 7.Uncertainty module.}
	\label{fig:detail}
\end{figure*}

%introduce our method
Figure \ref{fig:detail} shows the brief architecture of HGN framework. To summarize, we have following contributions:
\begin{itemize}
    \item [1)]
    We introduce an end-to-end ConvNets-Based gaze estimation framework called HybridGazeNet(HGN) which combines the ABMs and GBMs in a united framework which is the first trial in this area;
    \item[2)]
    We propose a hybridized training strategy to train HGN which could solve the misalignment problems between real data and synthetic data;
    \item[3)]
    Comprehensive experiments show that our HGN framework can improve the gaze estimation performance over within and cross datasets. Furthermore, the uncertainty module could learn the quality measurements which can be used under real application scenarios.
\end{itemize}

%-------------------------------------------------------------------------

\section{Related Work}
\subsection{Gaze estimation}
\label{sec:related}

Gaze estimation methods can be roughly divided into appearance-based methods(ABMs), geometric-based methods(GBMs)\cite{hansen2009eye} and ABM+GBMs.

\emph{\textbf{Geometric based methods}}. GBMs rely on features such as pupil center location, pupil center, and radius to estimate the eyeball model with a predefined geometric structure, and then inference the gaze direction\cite{venkateswarlu2003eye,alberto2014geometric,wood20163d,villanueva2013hybrid}. The common Pupil Center and Corneal Reflection (PC-CR) method\cite{guestrin2006general} always uses near-infrared light to produce glints on the eye cornea surface and captures images of the eye region, then gaze is estimated from the relative position between the pupil center and cornea. In \cite{jianfeng2014eye}, the authors set up a geometric model to calibrate the eyeball parameters via an RGB-D camera. \cite{wang2017real} proposes a 3D eye-face model which makes it feasible to estimate 3D gaze vector via web-camera. GBMs could achieve high accuracy but are sensitive to images resolution, and also require specific hardware.

\emph{\textbf{Appearance based methods}}. Recently, with the breakthrough of deep learning frameworks, ABMs have become popular in gaze estimation. By learning a map from eye images to gaze as a data-driven learning problem, ABMs are more robust to low-resolution images. In \cite{zhang2015appearance}, Zhang et.al release a large-scale gaze dataset called MPIIGaze and demonstrate that a shallow network can achieve high accuracy in gaze estimation. In \cite{zhang2017s}, the authors use full-face image as assistant input to estimate gaze direction. In\cite{cheng2018appearance}, Cheng et.al utilize two-eye asymmetry as a constraint to optimize the neural network. In\cite{cheng2020coarse}, the authors design a coarse-to-fine attention mechanism to capture subtle appearance features for gaze estimation. While the above works focus on searching for better appearance features for gaze estimation, another part of works solve the gaze estimation problem in the view of statistic learning. In\cite{xiong2019mixed}, Xiong et.al combine mixed-effects models with ConvNets to alleviate the negative effect of data distribution for existing gaze datasets. In \cite{wang2019generalizing}, the authors incorporate adversarial learning and Bayesian inference into a unified framework to improve generalization performance. However, as data-driven methods, ABMs can't capture all the variation in a certain image space with limited data and are prone to the over-fitting issue.

\emph{\textbf{ABM+GBM methods}}. ABM+GBM methods aim at learning gaze direction via combining ConvNets features and geometric informations, e.g. eye landmarks, head pose. In\cite{zhu2017monocular}, Zhu et.al designed a gaze transformer layer to connect separated head pose and eye movements model and the transformer layer can alleviate head-gaze correlation over-fitting. In\cite{yu2018deep}, the authors proposed a Constrained Landmark-Gaze Model(CLGM) to model the joint variation of eye landmark locations and gaze directions, aiming at making gaze estimator more robust and easier to learn. In\cite{Park_2018_ECCV}, the authors mapped 3D eyeball model into a 2D gazemap as external supervision for gaze estimation. In \cite{park2018learning}, the authors first detect 2D eye landmarks on image, and then use an average eyeball radius value from UnityEyes\cite{wood2016learning} to estimate gaze vector. Different from above study, our method take geometric model as part of neural network to guide the representation learning process.

\emph{\textbf{Quality Aware Gaze Estimation}}. Although the data quality issue is common in practical applications, there are not so many studies focusing on this problem in this area. In \cite{chen2019unsupervised}, 
authors proposed  manually designed training procedure to approach this problem. Despite the computational efficiency of the framework, the approach itself is hard to generalize to other scenarios. Different with this method, we focus on the eyeball structure and solve the problem of quality assignment in the view of statistic learning, which is more elegant and efficient.

%-------------------------------------------------------------------------RE
\subsection{Landmark detection} 
2D landmark detection technology is a well-established topic for many vision tasks, such as human pose estimation\cite{newell2016stacked}\cite{xiao2018simple}\cite{sun2019deep}, facial landmark detection\cite{wang2019adaptive}. Earlier methods \cite{sun2013deep}\cite{shi2016face} takes the image as the input and regress the coordinates directly. Recently, heatmap based methods achieve significant performance due to their ability to preserve spatial and contextual information. The landmark coordinates are predicted by finding the position in the heatmap with the largest response\cite{newell2016stacked}\cite{zhang2020distribution}. To eliminate such non-differentiable post-processing, the authors in [26] propose an integral operation on the heatmap and make the whole pipeline end-to-end for human pose estimation. In this work, we employ soft-argmax\cite{sun2018integral} mechanism for heatmap decoding so that the landmark location module is differentiable for the whole framework.

%-------------------------------------------------------------------------

%-------------------------------------------------------------------------
\section{Methods}
\begin{figure}[ht]
\centering
\includegraphics[width=8cm]{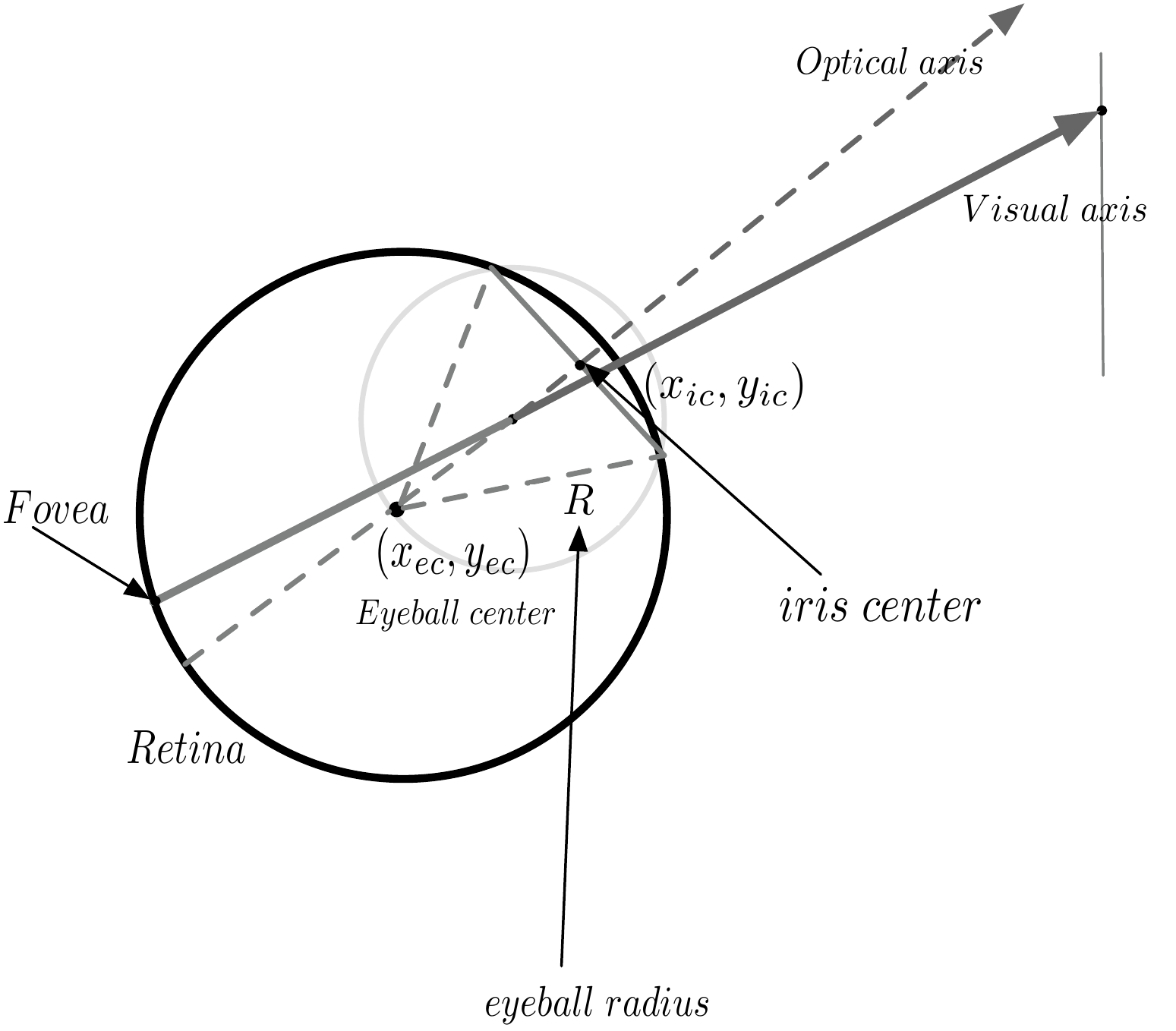}
\caption{3D eye model}
\label{fig:eye_ball}
\end{figure}

Figure \ref{fig:detail} shows the detailed architecture of our framework, HybridGazeNet takes single eye image as input and predicts gaze direction. The proposed method could summarized in two folds. Firstly, we design a HybridGazeNet framework, which encodes the gaze model into the ConvNets architecture; Secondly, we adopt a hybrid training strategy for HybridGazeNet.

\subsection{HybridGazeNet}
\label{hgn}
In order to encode geometric model into ConvNets, we need to dive into the geometric model first. A common 3D eyeball model is shown in Figure \ref{fig:eye_ball}, the posterior of the eyeball is called retina, and the center of the retina with the highest visual sensitivity is called the Fovea. The line joining the fovea with the center of corneal curvature is called the visual axis, and the connection between eyeball center and iris center is called optical axis. Specifically, the direction of visual axis is gaze direction. By simply regard the direction of optical axis as gaze direction, we can fit 3D eyeball model by knowing the eyeball radius, the 3D landmarks of eyeball center, and iris center location.
Suppose we have the 3D coordinate of following points: iris center, denote as $(x_{ic}, y_{ic}, z_{ic})$; eyeball center, denote as $(x_{ec}, y_{ec}, z_{ec})$; and eyeball radius, denote as $R$. And the gaze direction, denote as $\theta, \phi$. With these conditions, 2D iris center landmark location can be computed by following Eq. \ref{eq:2dlandmark}:
\begin{equation}
\label{eq:2dlandmark}
\left\{\begin{matrix}
\begin{aligned}
x_{ic} &= x_{ec} + Rsin(\phi)cos(\theta) \\ 
y_{ic} &= y_{ec} + Rsin(\theta) \\
\end{aligned}
\end{matrix}\right.
\end{equation}
Vice versa, with the iris center, eyeball center location and eyeball radius, gaze direction {$\theta$, $\phi$} can be solved with following Eq.\ref{eq:gazevector}. Denoting this mapping function as $Recon(\vec{x})$, where $\vec{x}$ represents 2D coordinate vector without $z_{ic}$ and $z_{ec}$. For clarification, all processes are computed in the camera coordinate system. We implement this reconstruction module at the top of ConvNets, as shown in Figure \ref{fig:detail}.
\begin{equation}
\label{eq:gazevector}
\left\{\begin{matrix}
\theta=arcsin(\frac{y_{ic}-y_{ec}}{R}) & \\ 
\phi = arcsin(\frac{x_{ic}-x_{ec}}{Rcos\theta}) & 
\end{matrix}\right.
\end{equation}

The eyeball radius branch predicts eyeball radius. Later in this section, we will introduce how to get radius ground truth from synthesis data. The loss term for eyeball radius is Equation.  \ref{radiusloss}:
\begin{equation}
\label{radiusloss}
L_r = ||\tilde{r}-r ||_{1}
\end{equation}
Coupled with landmark branch and eyeball radius branch, we propose a reconstruction module, which builds the relation between 2D landmarks and 3D gaze directions. A common 3D eyeball model is shown in Figure \ref{fig:eye_ball}, the posterior of the eyeball is called retina and the center of the retina with the highest visual sensitivity is called the Fovea, the line joining the fovea with the center of corneal curvature is called the visual axis, and the connection between eyeball center and iris center is called optical axis. Specifically, the direction of visual axis is gaze direction. By simply regard the direction of optical axis as gaze direction, we can fit 3D eyeball model by knowing the eyeball radius, the 3D landmarks of eyeball center, and iris center location.
Suppose we have the 3D coordinate of following points: iris center, denote as $(x_{ic}, y_{ic}, z_{ic})$; eyeball center, denote as $(x_{ec}, y_{ec}, z_{ec})$; and eyeball radius, denote as $R$; and gaze direction, denote as $\theta, \phi$. With these known conditions, iris center landmark location can be computed by following Equation: \ref{eq:2dlandmark}:
\begin{equation}
\label{eq:2dlandmark}
\left\{\begin{matrix}
\begin{aligned}
x_{ic} &= x_{ec} + Rsin(\phi)cos(\theta) \\ 
y_{ic} &= y_{ec} + Rsin(\theta)
\end{aligned}
\end{matrix}\right.
\end{equation}
Vice versa, with the iris center, eyeball center location and eyeball radius, the pitch $\theta$ and yaw $\phi$ of gaze direction can be computed as following Equation \ref{eq:gazevector}, denote this mapping function as $Recon(\vec{x})$ where $\vec{x}$ represents 2D coordinate vector:
\begin{equation}
\label{eq:gazevector}
\left\{\begin{matrix}
\theta=arcsin(\frac{y_{ic}-y_{ec}}{R}) & \\ 
\phi = arcsin(\frac{x_{ic}-x_{ec}}{Rcos\theta}) & 
\end{matrix}\right.
\end{equation}

From Eq. \ref{eq:gazevector}, we can find that knowing the 2D eye landmarks and eyeball radius, is able to estimate gaze directions. For $(x_{ec}, y_{ec})$ and $(x_{ic}, y_{ic})$, we can use the off-the-shelf landmark localization ConvNets. Different with conventional landmark localization tasks, in order to to make the whole framework differentiable, we adopt soft-argmax mechanism as heatmap decoding function. By doing so, the landmark detection branch can be trained in end-to-end manner. 

The soft-argmax mechanism can be described with following equation:
\begin{equation}
\label{eq:softmax}
        L_{i} = \sum_{p_y=1}^{h}\sum_{p_x=1}^{W}p\cdot \tilde{h_{i}}(p)
\end{equation}
Here, $H,W$ represent height and width of input image respectively, $\tilde{h_{i}}(p)$  is the normalized heatmaps for $i$-th heatmap:
\begin{equation}
    \tilde{h_{i}}(p)=\frac{e^{h_i(p)}}{\int_{q\in \Omega }^{}e^{h_i(p)}}
\end{equation}
For landmark location branch, the loss term is Eq. \ref{heatmaploss}:
\begin{equation}
\label{heatmaploss}
L_{h} = \sum_{i=1}^{10}\sum_{p}^{}||\tilde{{h}_{i}(p)}_{gt}-\tilde{h_{i}(p)}_{pred}||_{1}
\end{equation}
where 10 contains iris center, eyeball center and 8 iris landmarks.
After we get landmark location, we need eyeball radius parameter. In HGN, eyeball radius is predicted by ConvNets. The loss term for eyeball radius is described in Eq. \ref{radiusloss}:
\begin{equation}
\label{radiusloss}
L_r = ||\tilde{r}-r ||_{1}
\end{equation}
So far, coupled with landmark branch and eyeball radius branch, the geometric model is encoded into the ConvNets. We use the L1 loss(Eq. \ref{gazeloss}) as loss function for the predicted gaze direction $\vec{g} = {\theta, \phi}$, where:
\begin{equation}
\label{gazeloss}
L_{gaze} = ||\vec{g} - Recon(\vec{x})||_{1}
\end{equation}

After all, the total loss of our proposed architecture is:
\begin{equation}
\label{eq:finalloss_all}
	L_{total} = \beta_{1}L_h + \beta_{2}L_{r} + \beta_{3}L_{gaze}
\end{equation}
where $\beta_{1}, \beta_{2},\beta_{3}$ are weight coefficients, we set $\beta_{1}=5$, $\beta_{2}=1, \beta_{3}=1$ empirically. From Eq. \ref{eq:finalloss_all}, we can find that the whole framework is differentiable including the reconstruction module. With the end-to-end optimization ability, model base part and appearance-based part could benefit with each other.

\subsection{Hybridized Training Strategy}
\label{training}
%introduce hybrid training strategy
In order to train the HGN model, we need eye landmark locations and eyeball radius supervision. However, there is no public datasets providing eye landmark location and eyeball radius annotation. To solve this problem, we use UnityEyes to generate synthetic data as compensation data during training. UnityEyes could synthesize large amounts of eyeball images with a predefined 3D eye model and could provide comprehensive annotation for training. We use synthetic data to provide accurate landmark locations, and statistic Eyeball radius as supervision for the HGN model. Specifficly, eyeball radius is a mean value by fitting model-fit error in 10k UnityEye images\cite{park2018learning}. In the meanwhile, gaze supervision comes from both real and synthetic data. 
There are three issues we need to deal with during hybrid training. The first issue is coordinate system misalignment. As mentioned before, the location in Eq. \ref{eq:gazevector} is under camera coordinate system, and landmark locations generated by UnityEyes are under image coordinate system. Usually, we need to use camera intrinsic parameter to transform location from image coordinate to camera coordinate system. In this case, we assume that synthetic images is captured by a perfect camera which has no distortion. With this assumption, we don't need intrinsic parameter of UnityEyes for coordinate transformation. 

The second issue is the camera coordinate system misalignment. Though we don't use localization information from real data, we need to make sure that real data and synthetic data are under the same camera space. We use \emph{\textbf{Data Normalization}} technique to solve this misalignment. This part will clarified in Section \ref{datasets}. 

Last but not the least, there is a huge quality gap between synthetic data and real data. Gaze samples collection needs human machine collaboration, however, it's difficult to guarantee no mistake occurs during this procedure. As a result, the quality of real data is inferior then synthetic data. During training, the quality margin between these two type of data would degrade the optimization process. To address this problem, we introduce an \emph{\textbf{Uncertainty Module}} which would alleviate this negative impact.

Different with \cite{chen2019unsupervised}, we solve this problem in the view of statistic learning which is more effective and elegant. Same as \cite{he2019bounding}, We assume that the quality of collected data is distributed in a Gaussian distribution and also the label error. Following this assumption, we have Eq. \ref{eq:gaussian}, where $g_{e}$ and is the estimated gaze and $g$ is the ground truth. 
\begin{equation}
\begin{aligned}
	\label{eq:gaussian}
	f(x)=\frac{1}{\sqrt{{2\pi}}\times\sigma}e^{\frac{1}{2}(\frac{g - g_{e}}{\sigma})^2}
\end{aligned}
\end{equation}

Next, we apply KL-Divergence as loss function to supervise the output gaze of HGN. Applying Gaussian Distribution into KL-Divergence, for each sample we have Eq. \ref{eq:klloss}, in which $\sigma ^{2}$ represents the variance of each sample.
\begin{equation}
\begin{aligned}	
	\label{eq:klloss}
	L_{reg}&=D_{KL}(P_{D}(\vec{g})||P_{\theta}(\vec{g})) \\
	&= \int{P_{D}(\vec{g}) \log(P_{D}(\vec{g}))d\vec{g}} - \int{ P_{D}(\vec{\vec{g}}) \log(P_{\theta}(\vec{g}))d\vec{g}} \\
	&= \frac{(\vec{g} - \vec{g_{e}}) ^ {2}}{2 \sigma ^{2}} + \frac{\log \sigma ^{2}}{2} + \frac{\log (2 \pi)}{2} - H(P_{D}(\vec{g}))
\end{aligned}	
\end{equation}

Next, we apply KL-Divergence as loss function to supervise the output gaze of HGN. Applying Gaussian Distribution into KL-Divergence, for each sample we have Eq. \ref{eq:klloss}, in which $\sigma ^{2}$ represents the variance of each sample.
\begin{equation}
\begin{aligned}	
	\label{eq:klloss}
	L_{reg}&=D_{KL}(P_{D}(\vec{g})||P_{\theta}(\vec{g})) \\
	&= \int{P_{D}(\vec{g}) \log(P_{D}(\vec{g}))d\vec{g}} - \int{ P_{D}(\vec{\vec{g}}) \log(P_{\theta}(\vec{g}))d\vec{g}} \\
	&= \frac{(\vec{g} - \vec{g_{e}}) ^ {2}}{2 \sigma ^{2}} + \frac{\log \sigma ^{2}}{2} + \frac{\log (2 \pi)}{2} - H(P_{D}(\vec{g}))
\end{aligned}	
\end{equation}

Following the practice in \cite{he2019bounding}, the network predicts $\alpha = \log(\sigma ^ {2})$ instead, which is pictured in Figure \ref{fig:detail}. Finally we have the loss function of uncertainty module(Eq. \ref{eq:finalloss}), where $L_{g}$ is loss function for original gaze estimation.
\begin{equation}
	\label{eq:finalloss}
	\begin{aligned}
		L_{UM} = f(g_{e}) = e^{-\alpha}(L_{gaze} - \frac{1}{2}) + \frac{\alpha}{2}
	\end{aligned}
\end{equation}
Specifically, $\alpha \in \mathbb{R}^{2}$ is corresponding to pitch and yaw of gaze direction. From Eq. \ref{eq:finalloss}, we can find this term $e^{-\alpha}$ will control the loss scale for each sample, higher value leads to higher loss and the sample would have more impact on the network which means this sample is more valuable.

\section{Experiments}
In this section, we describe the experimental setup and evaluate the effectiveness of the HybridGazeNet on three public datasets: Columbia Gaze\cite{smith2013gaze}, UT Multiview\cite{sugano2014learning} and MPIIGaze\cite{zhang2015appearance}, following \textbf{\emph{within-dataset}} and \textbf{\emph{cross-dataset}} evaluation protocols in \cite{Zhang2020ETHXGaze}. 

%Section \ref{datasets} gives a brief introduction and pre-process procedure of these datasets. Next, in section \ref{ablation}, we decompose our methods with detailed ablation study in order to quantize the contribution of each module and analyze our uncertainty modeling with quantized metrics. After all, in section \ref{SOTA} we compare our method with state-of-the-art methods to show the effectiveness of our framework network.

\begin{table*}
	\begin{center}
		\begin{tabular}{l|ccccc}
		\toprule[1pt]
		\midrule[0.5pt]
			Method   & baseline & MTL & MTL-wo-radius &MTL-wo-lmks &HGN \\
			\hline Mean Error & 4.61 &4.23 &4.75 & 4.79& 3.70\\
		\bottomrule[1pt]
		\end{tabular}
	\end{center}
	\caption{MTL vs. HGN on Columbia Gaze}
	\label{tab:mtl}
\end{table*}

\subsection{Datasets}
\label{datasets}

\emph{\textbf{MPIIGaze}} contains 1500 left and right eye images of 15 participants, which were recorded under various conditions in head pose or illuminations and contains people with glasses. We use the provided normalized images for training, which are approximate of size 36×60 pixels, and are already frontalized relying on the head pose yaw and pitch. Unlike UT Multiview and Columbia collect gaze data under a carefully setting environment, MPIIGaze dataset is more close to daily life scenarios. 

\emph{\textbf{Columbia Gaze}} uses five high-resolution RGB-D cameras to record gaze samples, 5,880 samples with discrete gaze directions are collected, head pose of participants are fixed. We apply 5-folds cross- validation for within-dataset evaluation on Columbia.

\emph{\textbf{UT Multiview}} comprises 23040 (1280 real and 21760 synthesized) left and right eye samples for each of the 50 subjects. It was collected under strict laboratory condition, with various head pose. We use synthesized data provided by authors for experiments and apply 3-folds cross-validation for within-dataset evaluation on UT Multiview.

\emph{\textbf{MPIIGaze}} contains 1500 left and right eye images of 15 participants, which were recorded under various conditions in head pose or illuminations and contains people with glasses. We use the provided normalized images for training, which are approximate of size 36×60 pixels, and are already frontalized relying on the head pose yaw and pitch. Unlike UT Multiview and Columbia collect gaze data under a carefully setting environment, MPIIGaze dataset is more close to daily life scenarios.

\emph{\textbf{UnityEyes}} We use auxiliary database UnityEyes for acquiring eyeball model information. We generate 2000 synthetic samples w.r.t the gaze range for each dataset.

\emph{\textbf{Normalization}} Hybridized Training required images are normalized into a standard camera space. We normalized  the image and the head pose space as \ref{preprocessing}. After normalized, six degrees of freedom of object pose is normalized to  two degrees of freedom. Therefore, the appearance variation that needs to be handled inside the appearance-based estimation function has only two degrees of freedom as $(\theta,\phi)$, where $\theta$ and $\phi$ mean horizontal and vertical rotation angles respectively.

\subsection{Data preprocessing}
\label{preprocessing}
As mentioned before in section \ref{method}, HGN is based on single-eye image. Since Columbia Gaze, MPIIGaze, and UT Multiview provide data in different format, we adapt different preprocessing procedures for them. For UT Multiview, we directly use the synthesized eye image as training data. For Columbia dataset, we first detect face landmarks with our private landmark detection model and then norm the eye image with data normalization process. For MPIIGaze, we use the images as mentioned in \cite{zhang2017mpiigaze}, which has  been normalized already from the whole picture. And then, we employ histogram-equalized algorithm to the input eye images.

\emph{\textbf{Training details}}.
We choose resnet-18\cite{he2016deep} as our backbone network, and the input image size is 64 x 96. Specifically, our network is trained with ADAM for 100 epochs with the initial learning rate being 0.0001 and the batch-size is 64. The learning rate is reduced by a factor of 10 at epoch 20 and 60. Weight decay and momentum are set as 0.00005 and 0.9. We initialize our backbone networks with the weights pretrained on UnityEyes data. We employ data augmentation method by following settings in \cite{park2018learning} which contains image blurring, downscale-then-upscale, random brightness, random contrast, and additional lines for artificial occlusions.

\subsection{Ablation study}
\label{ablation}

\begin{figure}[t]
\centering
\includegraphics[width=8cm]{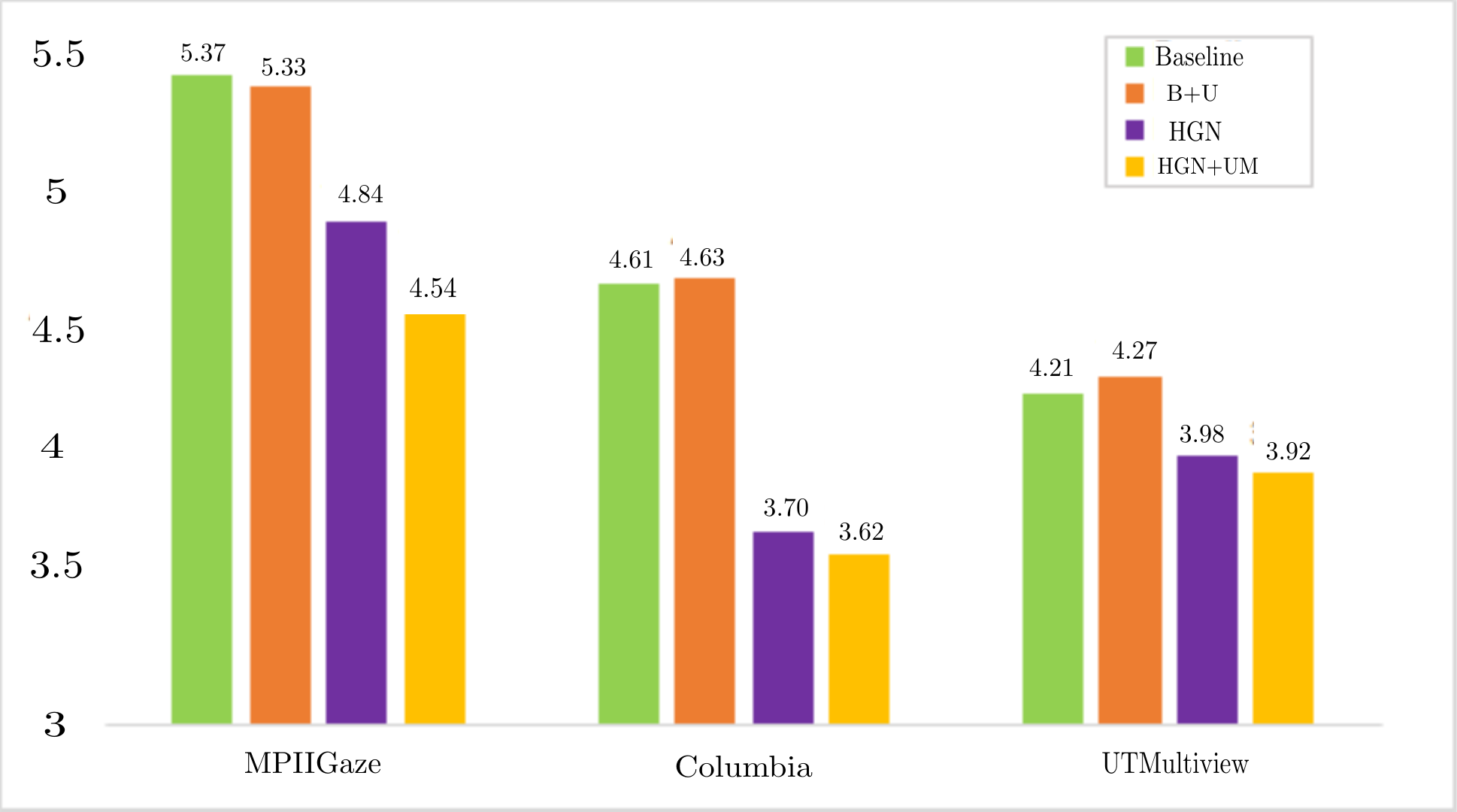}
\caption{Ablation study on three public datasets.}
\label{fig:ablation_study}
\end{figure}

\emph{\textbf{HybridGazeNet}} Firstly, we perform ablation study to demonstrate the effectiveness of HybridGazeNet. Considering following 4 models:
\begin{itemize}
	\item baseline(B): baseline ConvNets estimator, regress gaze direction directly;
\end{itemize} 
\begin{itemize}
	\item baseline+UnityEyes(B+U): baseline ConvNets estimator trained on real data and 2000 UnityEyes data;
\end{itemize}
\begin{itemize}
	\item baseline+UnityEyes+HGN(HGN): HybridGazeNet trained with synthetic data and real data;
\end{itemize}
\begin{itemize}
	\item baseline+UnityEyes+HGN + Uncertainty Modeling(HGN+UM): HybridGazeNet with uncertainty model.
\end{itemize}

\begin{figure*}[ht]
\centering
\includegraphics[width=13cm]{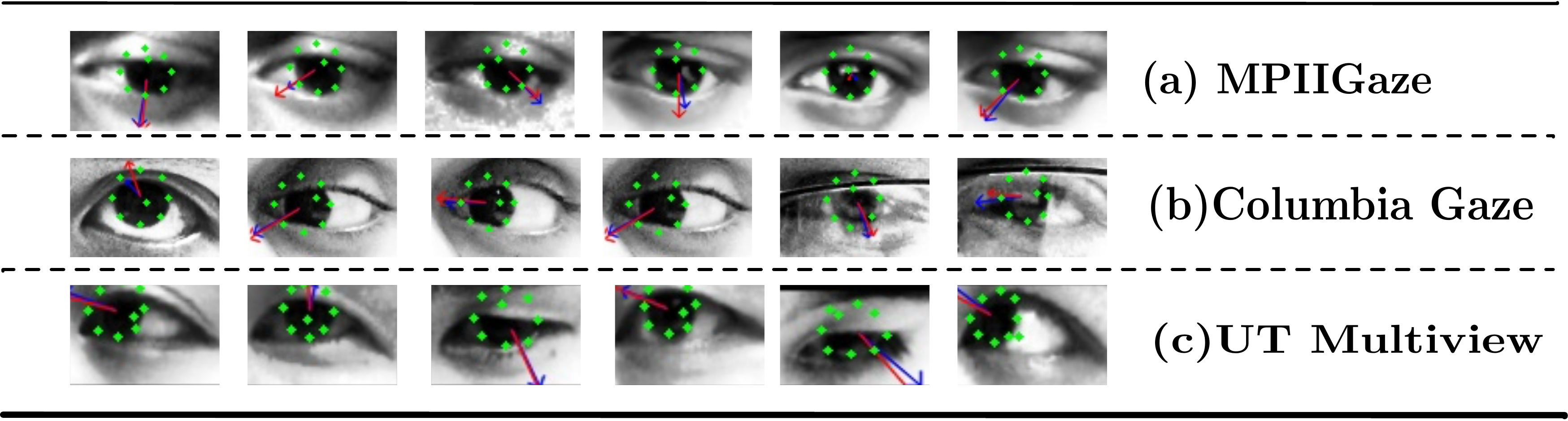}
\caption{Visualization of landmark detection results. Green dots are detected landmarks. Red arrows are predicted gaze direction while blue arrow are ground truth.}
\label{fig:result}
\end{figure*}

We conduct within-dataset ablation study on three public datasets. The results are presented in Figure \ref{fig:ablation_study}. Compare \textbf{B} and \textbf{B+U}, we can see that simply adding synthesis data into training set can not improve the gaze estimation performance. Especially, extra synthesis data decrease the gaze estimation performance on UT Multiview dataset. With HGN, angle error decreases by $0.53^\circ$, $0.91^\circ$, $0.23^\circ$ on three datasets. This indicates that with geometric model, HGN learns better eye feature for gaze estimation. Figure \ref{fig:result} shows the visualized predicting results on different datasets. Although the geometric structure supervision is from the synthetic data, with the hybrid training strategy HGN can perform well on eye landmark location and achieve outstanding gaze estimation performance. Another important part is uncertainty modeling. As shown in \textbf{HGN + UM}, with this module, angle error decreases by $0.83^\circ$, $0.99^\circ$, $0.29^\circ$ compared with baseline method. Besides the improvements contributed by this module, we will give another detailed analysis of this module in the next paragraph. In addition, our HGN and UM  don't increase inference time. 

\begin{figure}[ht]
\centering
\includegraphics[width=9cm]{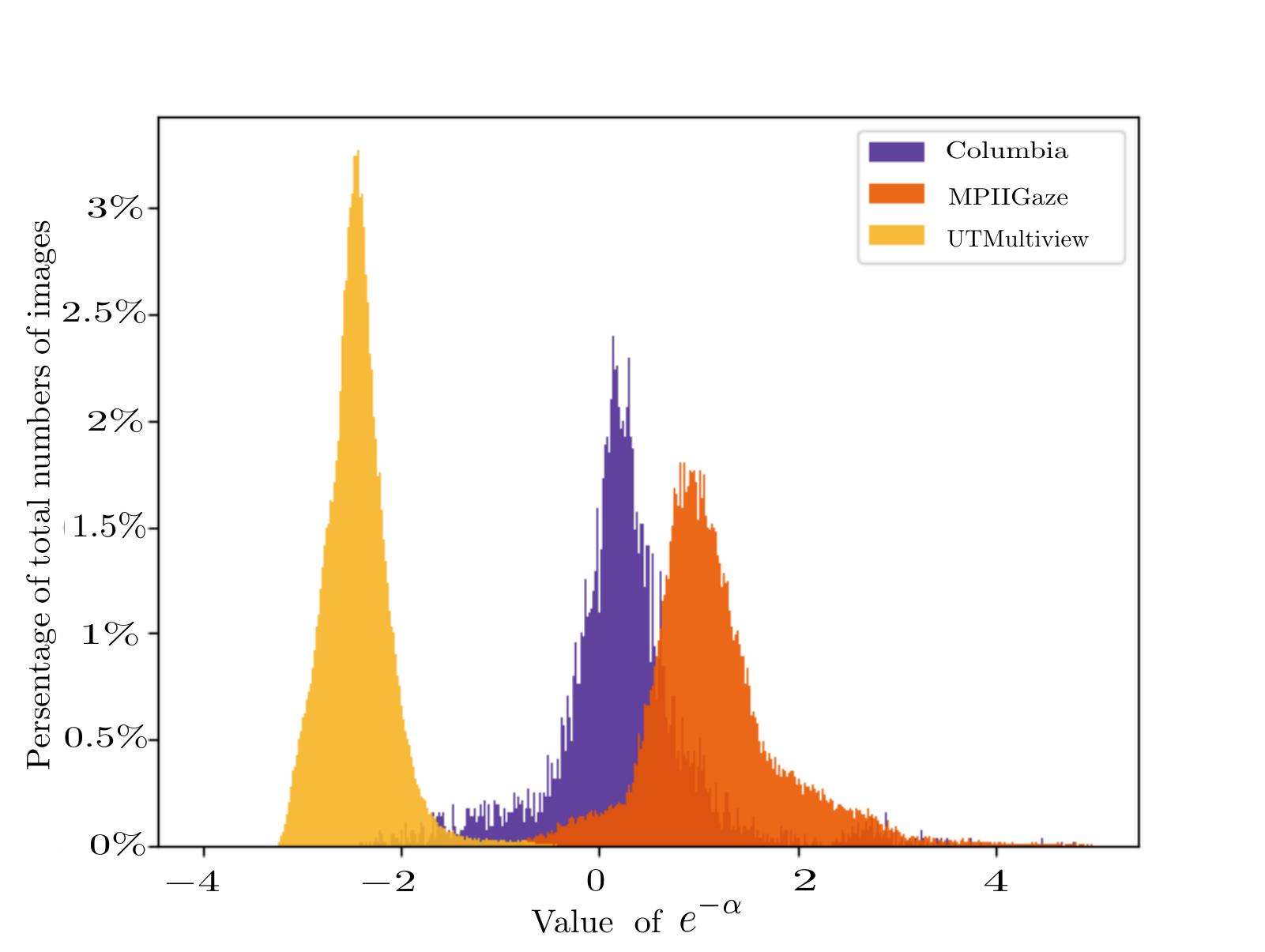}
\caption{Distribution of $e^{-\alpha}$ on different gaze dataset.}
\label{fig:sigma}
\end{figure}

\begin{figure*}[t]
\centering
\includegraphics[width=16cm]{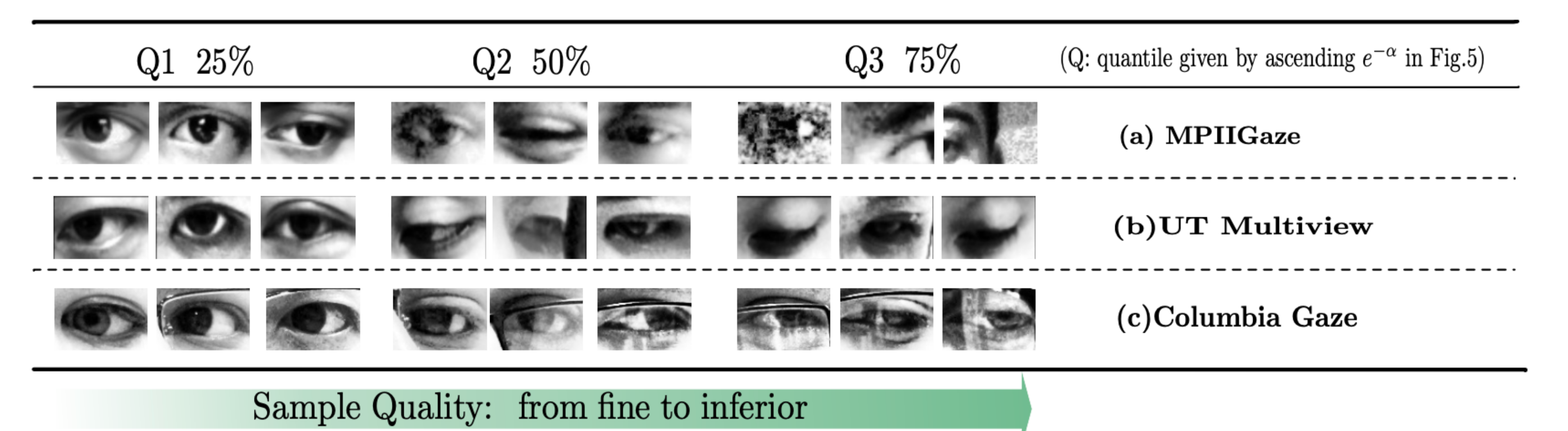}
\caption{Eye images with different sigma value on MPIIGaze, Columbia Gaze, and UT Multiview dataset.}
\label{fig:sigma_sample}
\end{figure*}

\emph{\textbf{Reconstruction Module}}. To demonstrate the performance improvement comes from  reconstruction module, we implement a multi-task learning network as which predicts eye radius, eye landmarks and gaze direction jointly. Unlike reconstruction module integrating eye radius and eye landmarks to reconstructing gaze, MTL predict the three labels by splitting whole network into three separated branches. We also remove the uncertainty branch. The comparison results between Multi-task learning(MTL) and HGN is show in Table \ref{tab:mtl}, \emph{MTL-wo-radius} and \emph{MTL-wo-lmks} refer to MTL is without radius branch or landmarks branch, respectively. From Tabel \ref{tab:mtl} we can see that MTL achieves $4.23^{\circ}$, outperforms baseline by $0.4^{\circ}$, however is still inferior than HGN. This demonstrate that with the geometric model, ConvNets could achieve better accuracy for gaze estimation.

\emph{\textbf{Uncertainty Module}}. In this part, we will further analyze the numerical distribution of learned $\alpha$ on each dataset. As stated in section \ref{training}, we can view $e^{-\alpha}$ as the sample quality. Figure \ref{fig:sigma} shows the statistical results of $e^{-\alpha}$ on each dataset. From left to right are UT Multiview, Columbia, and MPIIGaze. This indicates that UT Multiview has the best quality of labeled samples and MPIIGaze dataset is relatively inferior. Recall the data collection process of each dataset described in section \ref{datasets}, we can find that under the daily life scenario collecting precise gaze samples is quite challenging. In Figure \ref{fig:ablation_study}, the \textbf{HGN + UM} achieves the best result, indicates that with the uncertainty module, the performance of gaze estimation would be better. Moreover, the improvement in MPIIGaze dataset is larger than other two datasets, which means the UM module is effective to alleviate the negative effect of noise data. Besides the analysis, we sort the samples with $e^{-\alpha}$ and visualize samples over different quantiles. We can see that fine samples has clear eyeball structure, while the inferior samples lost the information of eye region.

\begin{table}
	\begin{center}
		\begin{tabular}{l|cccc}
		\toprule[1pt]
		\midrule[0.5pt]
			Method   &Columbia   &MPIIGaze   &UT Multiview \\
			\hline CLGM\cite{yu2018deep}     &-        &-        &5.7\\
			ELG\cite{park2018learning}        & 6.2    &4.6    & 11.5\\
			DPG\cite{Park_2018_ECCV}      &3.8        &4.5        &-\\
			MeNet\cite{xiong2019mixed}   &-           &4.9        & 5.5\\
			ours &\textbf{3.7}        &\textbf{4.5}        &\textbf{3.9}\\
			\hline
		\bottomrule[1pt]
		\end{tabular}
	\end{center}
	\caption{Within-dataset Evaluation}
	\label{tab:within}
\end{table}

\begin{table}
\begin{center}
\begin{tabular}{l|ccc}
\toprule[1pt]
\midrule[0.5pt]
Method   &Columbia   &MPIIGaze  \\
\hline MPIIGaze\cite{zhang2017mpiigaze} &-          &9.9 \\
ELG\cite{park2018learning}      &8.7        &8.3 \\
MeNets\cite{xiong2019mixed}   &-          &9.5 \\
Bayesian\cite{chen2019unsupervised} & $\approx${7.8}   &\textbf{7.4} \\
ours     &\textbf{6.2}        &8.9 \\
\hline
\bottomrule[1pt]
\end{tabular}
\end{center}
\caption{Cross-dataset Evaluations}
\label{tab:cross}
\end{table}

\subsection{Comparison with State-of-the-art}
\label{SOTA}
After the detailed ablation study, we compare our proposed method with state-of-the-art methods on three public datasets. We choose four methods as compared methods, which are CLGM\cite{yu2018deep}, DPG\cite{Park_2018_ECCV}, ELG\cite{park2018learning}, and MeNets\cite{xiong2019mixed}. Tabel \ref{tab:within} shows within-dataset evaluation results. They all use single eye image as model input and use the same evaluation protocol as our work. On Columbia Gaze, we outperform the SOTA method DPG\cite{Park_2018_ECCV} by 0.1 degrees. On UT Multiview, our algorithm(3.9 degrees) outperforms the state-of-the-art(5.4 degrees) by a large margin(27.7$\%$ improvement). Our proposed HGN gets limits improvement on MPIIGaze, as explained in Section \ref{ablation} the collection process of the datasets is different from another two datasets which have lots of inferior samples. With further analysis, we find that MPIIGaze exists more noise samples than Columbia Gaze and UTMultiview, the corrupted sample is shown in Figure \ref{fig:sigma_sample}, most of the inferior sample are due to eye blinks, blurring, or false landmark detection.   

For cross-dataset evaluation, we train on UT Multiview then test on Columbia Gaze and MPIIGaze. The cross-dataset experiment results are shown in Tabel \ref{tab:cross}. Our approach achieves a larger improvement of 1.7$^\circ$ degrees over state-of-the-art on Columbia and better than most existing methods on MPIIGaze. Same as within-dataset result, HGN performs better on high-quality dataset. 

\section{Conclusion}
To our best knowledge, this is the first attempt to encode an explicit eyeball-gaze model and appearance-based methods into a unified framework for gaze estimation. We introduce a novel network named HGN(HybridGazeNet) which can be trained end-to-end rely on mixing real and synthesis data. HGN encode geometric constraint which can 1) transfer geometrical information from synthesis data into real data to promote learning robust feature for gaze estimation. 2) narrow the gap of different datasets, and make full use of annotations that lead to better training cross datasets. Besides, we also proposed an uncertainty model which can quantize the quality of gaze sample and alleviate the influence of noise sample during HGN training. This is quite meaningful for practical application, quantized quality value could dropout noise labeled data and improve robustness for gaze estimation algorithm.  

In the future, we can consider that introduce more complicated eyeball model into our framework, by integrating more feature, the method will be more robust. In the meantime, we will focus on self-surpervised learning for gaze estimation to further improve the performance under unconstraint environment.

%-------------------------------------------------------------------------

%-------------------------------------------------------------------------

%-------------------------------------------------------------------------

%------------------------------------------------------------------------

{\small
\bibliographystyle{ieee_fullname}
\bibliography{egbib}
}

\end{document}